# DCTNet : A Simple Learning-free Approach for Face Recognition

Cong Jie Ng[*] and Andrew Beng Jin Teoh[†]
Yonsei University, Seoul, South Korea
E-mail: congjie@yonsei.ac.kr[*], bjteoh@yonsei.ac.kr[†]

*Abstract*— **PCANet was proposed as a lightweight deep learning network that mainly leverages Principal Component Analysis (PCA) to learn multistage filter banks followed by binarization and block-wise histograming. PCANet was shown worked surprisingly well in various image classification tasks. However, PCANet is data-dependence hence inflexible. In this paper, we proposed a data-independence network, dubbed DCTNet for face recognition in which we adopt Discrete Cosine Transform (DCT) as filter banks in place of PCA. This is motivated by the fact that 2D DCT basis is indeed a good approximation for high ranked eigenvectors of PCA. Both 2D DCT and PCA resemble a kind of modulated sine-wave patterns, which can be perceived as a bandpass filter bank. DCTNet is free from learning as 2D DCT bases can be computed in advance. Besides that, we also proposed an effective method to regulate the block-wise histogram feature vector of DCTNet for robustness. It is shown to provide surprising performance boost when the probe image is considerably different in appearance from the gallery image. We evaluate the performance of DCTNet extensively on a number of benchmark face databases and being able to achieve on par with or often better accuracy performance than PCANet.**

## I. INTRODUCTION

Deep convolutional network shows its success in various image classification tasks has drawn significant attention in recent years [1][2][3]. The key ingredient of the success is the ability to automatically discover and learn abstract representation of the data build up in multiple stages where each stage represents intermediate level representation developed from the previous stage. Nonetheless, besides filter learning, one of the key challenges is designing the proper network architecture and choosing the right configuration and parameters such as number of layers, filter size, choice of pooling function and etc. AlexNet [1], which outperformed the runner up by 10% error gap in 2012 ILSVRC challenge [4] adopts similar architecture as early convolution network, ie. LeNet [5], but with deeper and bigger network structure. GoogLeNet [2] adopts Inception module inspired by Network in Network [6] won ILSVRC 2014 [4].

Despite the successes, the feature learning mechanism and optimal network configurations of deep networks are not well understood [7]. Scattering Convolution Network (ScatNet) [7] that is based on scattering theory addresses these open problems partially. With prefixed filters generated from mathematical functions, ScatNet demonstrates state-of-the-art performance over ConvNet [8] in handwritten recognition and texture discrimination tasks.

Recently, a lightweight unsupervised deep learning network proposed by Chan et al. called PCANet (Principal Component Analysis Network) [9] works unexpectedly well in most of the image classification tasks despite very simple architecture. PCANet processes an input image via a layer-wise convolution with PCA filters and followed by binarization, block-wise histograming and eventually yield a long histogram feature vector. The histogram vector can be further compressed via dimension reduction technique such as whitening PCA. Prior to PCANet, a similar filter called Binarized Statistical Image Features (BSIF) [10] is proposed. BSIF binarizes the filter responses obtained from the convolution of an image with Independent Component Analysis (ICA) learned filters. However, BSIF is merely treated as an image descriptor in [14] but not expanded to a network form.

In this paper, we propose a much simpler learning-free alternative of PCANet via 2D DCT filters dubbed DCTNet, specifically tailored for face recognition. The choice of 2D DCT basis as filter bank is inspired by the Karhunen Loève Transform (KLT) in transform coding literature, which is also known as Principal Component Analysis (PCA) in multivariate statistics community. KLT is an optimal orthogonal transform that can decorrelate any signal completely and condense the signal energy maximally. However, despite these attractive properties, 2D DCT is chosen as the baseline JPEG image compression standard instead due to the reasons that KLT is data dependence and there is no fast algorithm available for KLT, which requires $O(N^3)$ to solve eigenvalue problem of the $N \times N$ dimension covariance matrix, whereas 2D DCT can be computed with $O(N \log_2 N)$ operations [11]. Apart from low complexity, 2D DCT computation is independent from data, which implies learning-free. Both 2D DCT and PCA filters are indeed equivalence when an image is assumed to be the first order Markov process subject to the condition when the local correlation between neighborhood pixels is high. We will elaborate this interesting fact in section III.

On the other hand, block-wise histograming of PCANet that is capable of implicitly encoding spatial information of image regions is useful for classification task like face recognition [12]. Block-wise histograming is essentially used to estimate the probability distribution function (pdf) of the image features in block-wise manner. Bigger block size implies better pdf estimation due to larger number of feature samples but poorer spatial precision. However, small block size introduces another problem in which the number of histogram bins would be more than the number of samples, and hence the resulting histogram becomes very sparse. Sparse histogram may render poor pdf estimation for that particular block.

In order to mitigate this trade-off, we propose an effective method called Tied Rank Normalization (TR Normalization) to regulate the histogram of DCTNet for robustness despite under sampling. Our proposed technique is based on tied rank principle inspired by Spearman's rank correlation [13] that computes the Pearson correlation between ranked variables. By adopting the ranking idea that is well tolerated to outliers, the appearance disparity between probe and gallery samples due to pose variation and occlusion, can be eliminated and hence to provide better robustness. In addition to that, we also adopt the intra-normalization proposed by [14] to spread the concentrated component energy of histogram vector more evenly, which was shown to be beneficial for accuracy performance improvement.

In a nutshell, the contribution of this paper is three-fold:
- We propose a much simple learning-free DCTNet by adopting 2D DCT bases as filter bank, which was shown equivalence to PCANet subject to certain condition.
- We also propose a histogram normalization technique called Tied Rank Normalization (TR Normalization) to eliminate the disparity of histogram vector of DCTNet based on the tied rank principle used by robust statistic (Spearman's rank correlation) and intra-normalization for feature-evenization.
- Lastly, we provide extensive experiment results with a number of benchmark face datasets for the proposed learning-free DCTNet. The datasets considered, ie. AR, FERET-I ('b' subset) and FERET-II ('fa', 'fb', 'fc', 'dup I', 'dup II' subset), covers various undesirable scenarios in face recognition such as variations in poses, lighting, expression, occlusion and time span.

## II. PRELIMINARY

PCANet [9] is designed to be a lightweight convolutional neural network (CNN) in which the filters in the convolution layers are learned by PCA, an unsupervised learning method as opposed to supervised learning approach via Backpropagation adopted in CNN. Unlike typical CNN, this simplistic CNN has no nonlinear operation in between layers instead the operation is only performed at the output layer. The nonlinear operation in PCANet refers to the binary thresholding operation that converts the filter responses into a binary map. Then, block-wise histograming is carried out to encode the spatial relation between blocks [12]. The detail of binarization and block-wise histograming is given in Section IV(B). Finally, the output feature vector is formed by concatenating all block-wise histograms.

Despite unsupervised, Chan et al. [9] also shows that by replacing the PCA filters with the linear discriminant analysis (LDA) learned filters, which exploits the class labels does not offer significant advantage over PCANet.

## III. METHODOLOGY

### A. Relationship Between DCT and PCA

In this section, we present a theoretical and empirical justification on the equivalence of PCA and DCT [11]. In essence, the local correlation between neighborhood pixels of an image makes it convenient to be regarded as a stochastic process, which can be modeled by a two dimensional stationary first order Markov process.

Without loss of generality, given a 1D signal with samples $\{x_i| i=1,...,N\}$, the correlation between any two samples $x_i$ and $x_j$ is defined as $r^{|j-i|}$ where $0 \leq r \leq 1$. $r^{|j-i|}$ indicates the correlation between two samples that declines exponentially as they get further apart. With the definition, the correlation matrix of this Markov chain is defined with a Toeplitz matrix as follows with all diagonal elements being the same as 1 has the highest correlation value [11].

$$R = \begin{bmatrix} 1 & r & r^2 & \dots & r^{N-1} \\ r & 1 & r & \dots & r^{N-2} \\ r^2 & r & 1 & \dots & r^{N-3} \\ \vdots & \vdots & \vdots & \ddots & \vdots \\ r^{N-1} & r^{N-2} & r^{N-3} & \dots & 1 \end{bmatrix} \quad (1)$$

It is shown by [15] that the principal components (eigenvectors) and its associated variances (eigenvalues) in eq. (1) can be obtained by performing eigen-decomposition on $R$. Hence, the $n^{th}$ eigenvalue is given as

$$\lambda_n = \frac{1-r}{1 - 2r\, cos\omega_n + r^2} \quad (2)$$

and the $m^{th}$ element of $n^{th}$ eigenvector is given as

$$\phi_{mn} = \left(\frac{2}{N + \lambda_n}\right)^{\frac{1}{2}} \sin\left(\omega_n\left(m - \frac{N-1}{2}\right) + \frac{(n+1)\pi}{2}\right) \quad (3)$$

where $0 \leq m, n \leq N - 1$ and $\omega_n$ is the $N$ real roots of the following equation

$$\tan(N\omega) = -\frac{(1-r^2)sin\omega}{(1+r^2)\cos\omega - 2r} \quad (4)$$

The relationship of PCA and DCT is unveiled when $r$ approaches 1 [11]. The following equation shows that the eigenvector of the resulting eigendecomposition of $R$ is indeed identical to DCT bases for $n > 0$ has $\lambda_n = 0$ and $\omega_n = n\pi/N$ when $r \rightarrow 1$:

$$\phi_{mn} = \sqrt{\frac{2}{N}} \cos\left(\frac{n\pi}{2N}(2m + 1)\right) \quad (5)$$

For the case when $n = 0$, it takes the following form which has $\lambda_0 = n$ and $\omega_0 = 0$

$$\phi_{m0} = \sqrt{\frac{1}{N}} \quad (6)$$

where $0 \leq m < N-1, 1 \leq n \leq N-1$

In summary, given $M$ overlapping blocks of a signal with size $N$ of stride 1, if the correlation between blocks is very high, the PCA eigenvector of the blocks covariance matrix will approach DCT basis. It is also worth noting that when $r = 1$, the eigenvectors are no longer unique as all the elements of the correlation matrix become unity, which imply singularity.

To gain further insight into the relationship of DCT and PCA, eq. (2) is plotted with $r = 0.9$, $N = 100$ and $\omega_n = \frac{n\pi}{N}$ as shown in **Fig. 1**. The plot suggests that large eigenvalue of PCA corresponds to low frequency in DCT and vice versa. This property is vital for DCT basis selection for DCTNet in section V, which follows the PCA by ranking the importance of eigenvector based on the respective eigenvalue. This property also explains the reason why zig-zag scanning is adopted in Baseline JPEG.

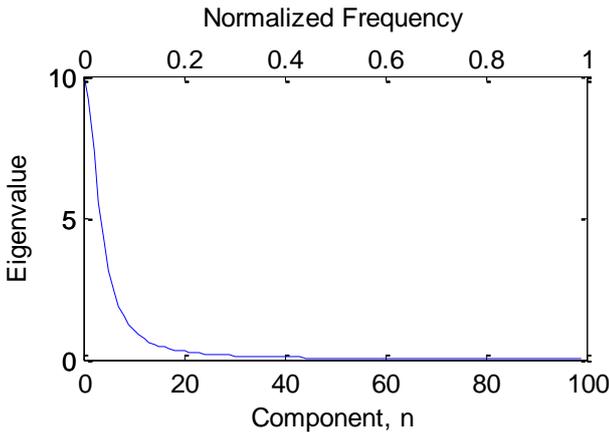

Fig. 1 Plot of (2) shows the inverse exponential relationship between eigenvalue and frequency

### B. 2D DCT and PCA

Albeit the derived equations show the equivalence of 1D DCT and PCA when $r$ approaches 1 in stationary first order Markov process model, it is shown to be applicable to 2D DCT on image too. Without resort to the rigorous proof that requires decomposing a much more complicated Toeplitz Matrix than (1), the similarity of 2D DCT and PCA eigenvectors is shown pictorially. To generate PCA bases, we use gray-scale frontal faces with expression and illumination of FERET 'b' subset dataset ('ba', 'bj' and 'bk') [16], which composed of 600 images of size $64 \times 64$ each. Each image is first segmented into overlapping patches of size $5 \times 5$ with stride 1. Each extracted patch is then vectorized into a 25 dimension vector. Lastly, eigen-decomposition is performed on the vectorized patches covariance to obtain the eigenvectors. To show the similarity between 2D DCT and PCA, the eigenvectors are reordered manually to be alike with 2D DCT basis ordering for better visualization as shown in **Fig. 2**.

Note that the eigenvectors may look quite different from the corresponding 2D DCT basis due to negation in the numeric sign. Besides sign inversion, both 2D DCT basis and PCA learned eigenvector from FERET 'b' subset are shown to have very similar structure. These bases can be essentially perceived as a filter bank with different cutoff frequencies at horizontal direction, vertical direction and their products.

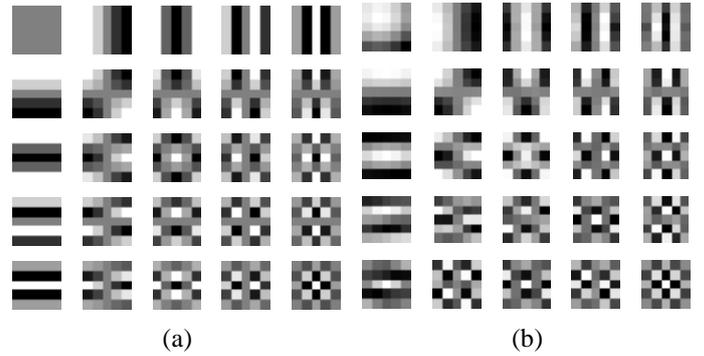

(a)          (b)

Fig. 2 (a) shows the DCT bases, (b) shows the PCA learned eigenvectors on FERET ba, bj and bk dataset with manual reordering for illustration purposes.

### IV. DISCRETE COSINE TRANSFORM NETWORK (DCTNET) ARCHITECTURE

DCTNet adopts a similar structure to PCANet except there is an extra layer at the histogram output for histogram normalization as shown in **Fig. 3**. The detail of each component is described below.

### A. Convolution Layer

Assume that filter size of all stages have the same size $k \times k$. Given an input image $I_d$ of size $m \times n$ with $D$ channels (multiple channel image or input from previous layer), boundary of each channel $d$ is zero padded with pad size $(k-1)/2$ before convolution to keep the size of output $O_d^p$ same as $I_d$. With a set of 2D DCT bases selected as described in section V denoted by $W_p^l \in \mathbb{R}^{k \times k}, p = 1,2,\ldots,P_l$ where $P_l$ is the number of filters at layer $l$, convolving each with $I_d$ yields

$$O_d^p = \{I_d * W_p^l\}_{p=1}^{P_l} \quad (7)$$

The number of output of each layer is $d.P_l$. Cascading this layer can form a deeper network. Since, there is no nonlinear operation in between the previous convolution layer and the next layer, DCT bases of each layer can be combined to form a flat single layer network. The number of bases formed is $\prod_{i=1}^{L} P_i$ where $L$ represents the number of convolution layers. For the sake of convenience without storing large number of combined filters and to ease the binarization process, the flat single layer architecture is not considered in this paper.

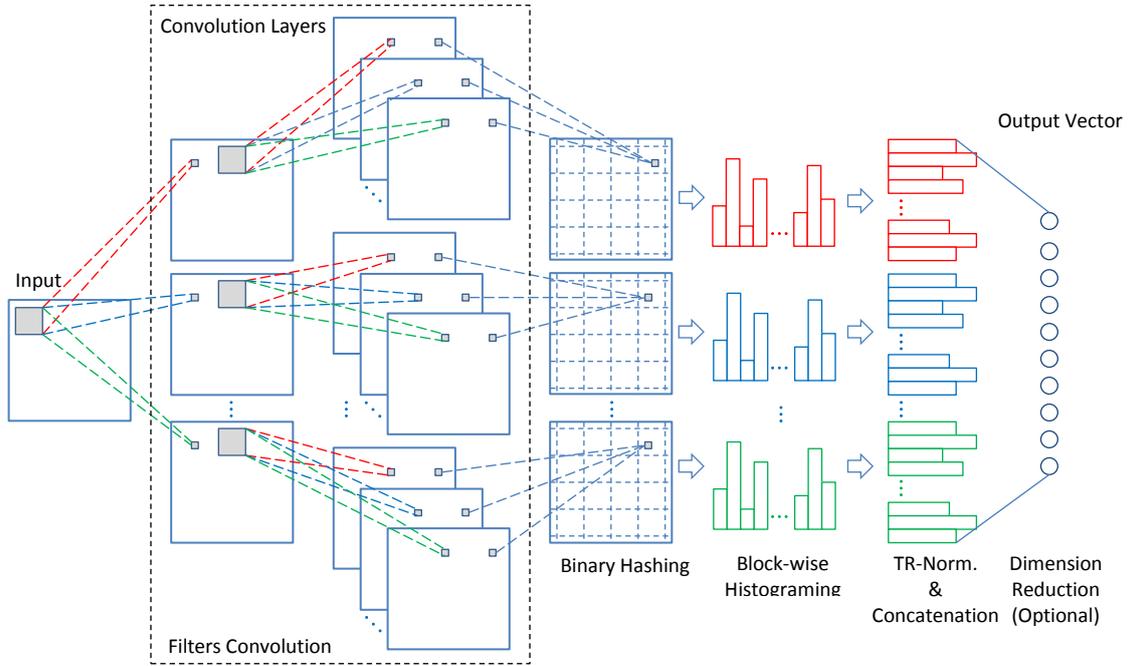

Fig. 3 The block diagram of the proposed DCTNet

*B. Binarization and Block-wise Histograming*

The last convolution layer of DCTNet forms $D$ sets of real valued outputs. Each set has a total of $P_L$ outputs where the outputs are the response of DCT filters. Binarization is performed on each set separately by first binarizing the responses with threshold at zero (value one for positive response, zero otherwise) denoted by $BIN(.)$. Followed by binarization, each binary string is encoded as a single integer number $\sum_{p}^{P_L} 2^{p-1} BIN(O_d^p)$ and forming an "image" for each set of $d^{th}$ output where each pixel has an integer range of $[0, 2^{P_L}-1]$. Then, each of these $D$ binarized "image" is partitioned into $B$ non-overlapping blocks. Histogram of each block denotes by $H_b^d$, $b = 1,2,...,B$; $d = 1,2,...,D$ with bin $[0, 2^{P_L}-1]$ is obtained as the input for histogram normalization layer that will be described in next section.

It is also worth to mention that block-wise histogram not only encodes spatial information [12], it also provides local translation invariance in the extracted features within each blocks. The combination of binarization and block-wise histograming is expected to be able to extract discriminative features.

*C. Histogram Tied Rank Normalization (TR Normalization)*

The first stage of TR normalization uses tied rank principle that computes rank of a given vector $x$ which produces a vector $\bar{x}$ that has a range from 1 to the length of $x$ where each element $\bar{x}_i$ corresponds to the ascending order rank of $x_i$. In case of ties, their average rank is assigned to all ties which may produce non-integer values. Given $H$ as the extracted block-wise histogram of a given face data, where $H = \{H_b^d\}_{b=1,d=1}^{B,D}$. Each $H_b^d$ is ranked with tied ranking without considering the bin with zero occurrence denoted by $\bar{H}_b^d$. This is because bin with zero occurrences is not a sample in histogram, it should be ignored in the ranking process. In order to make $\bar{H}_b^d$ to be more evenly distributed, we first apply square root on $\bar{H}_b^d$ forming $v_b^d = \sqrt{\bar{H}_b^d}$. Follow by L2 norm normalization which follows the idea of intra-normalization uses by [14] we obtain $\hat{v}_b^d$. The final TR normalized histogram feature vector is constructed by concatenating all $\hat{v}_b^d$

$$v = [\hat{v}_1^1, \hat{v}_2^1, ..., \hat{v}_B^1, \hat{v}_1^2, ..., \hat{v}_B^D] \in \mathbb{R}^{(2^{P_L})B\,D} \quad (8)$$

| Algorithm 1 : Histogram TR Normalization |
|---|
| **Input:** |
| Extracted block-wise histogram of an image : $H$ |
| **Output:** |
| TR normalized histogram feature vector : $v$ |
| **Start:** |
| 1. For each $H_d^b$ compute tied rank without bin with zero occurrence yields $\bar{H}_d^b$ |
| 2. $v_b^d = \sqrt{\bar{H}_b^d}$ |
| 3. Normalize $v_b^d$ with L2 norm to obtain $\hat{v}_b^d$ |
| 4. Repeat step 1 to step 3 for $b = 1,2,...,B$; $d = 1,2,...,D$ |
| 5. Concatenate all $v_b^d$ to obtain the final output $v$ |

The pseudo code of histogram TR normalization is shown in Algorithm 1. The TR normalized block-wise histogram is shown in **Fig. 4**. The disparity of the original block-wise histogram is shown to be eliminated and it is also shown to be more evenly distributed. Finally, the dimension of the resulting TR normalized block-wise histogram vector is optionally compressed with whitening PCA (WPCA) to obtain the final feature vector where the projection matrix is learned from Gallery set.

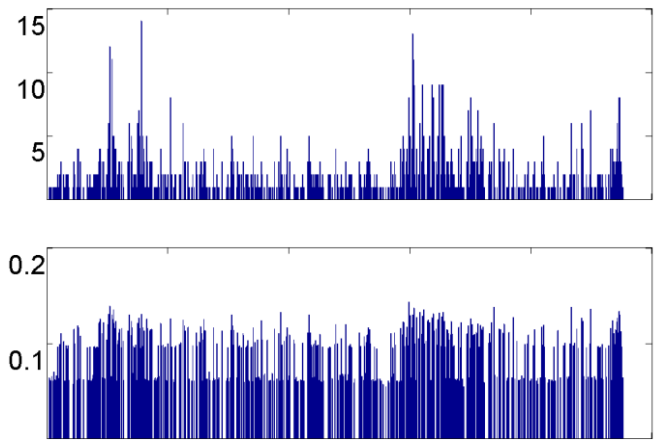

Fig. 4 Top shows a part of the original block-wise histogram feature vector; bottom shows the resulting TR normalized block-wise histogram feature vector. Note that, scale difference between the input and the output is due to normalization process

## V. SELECTION OF DCT BASES AS FILTER BANK

One essential issue to address when adopting 2D DCT basis into the network as filter bank is the basis selection. Unlike PCANet, eigenvectors are ranked by their respective eigenvalue strength. The first $P$ eigenvectors with the highest eigenvalue are selected as the network filter for each level. To address the issue one can refer to the derived eigenvalue equation (2) and **Fig. 1** as discussed in previous section which shows that eigenvalue has inverse exponential relationship. Low frequency DCT basis corresponds to high ranked eigenvector. Although (2) corresponds to 1D DCT, 2D DCT is just a product of vertical basis and horizontal basis of 1D DCT. Without lengthy mathematical proof for simplicity one can assume that the eigenvalue of the horizontal basis, vertical basis and the diagonal bases of the same frequency have the same value. That is to say, bases in the same antidiagonal row as shown in **Fig. 5** are assumed to have the same eigenvalue hence they are ranked equally.

To further rank the equally ranked DCT bases, the prior-knowledge of human face characteristic is taken into account. Since human face distinct features are composed of more high frequency horizontal components (eyes, eyebrows and lips) than low frequency vertical component, it is natural to rank the 2D DCT bases by horizontal-frequency major order. As illustrated in **Fig. 5**, zig-zag scanning used by Baseline JPEG alternates the frequency direction importance at each turn; the DCTNet keeps the importance of horizontal frequency direction at each turn to extract a more representative face features.

Lastly, DC component is not considered as a filter in DCTNet as reported by PCANet removing mean of each patch yields better performance. The basis selection is therefore starting from 2 to $P + 1$ in the horizontal-frequency major scanning order. Omitting the DC component which extracts the lowest frequency component or mean of the patch can improve the robustness of the extracted feature against global illumination changes.

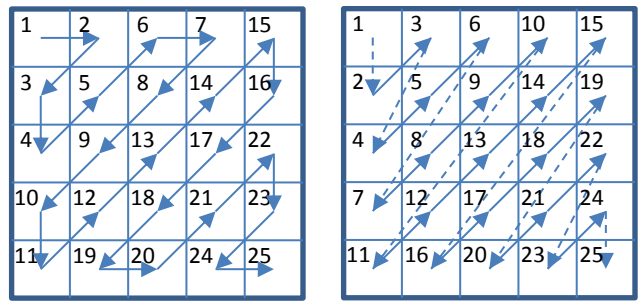

Fig. 5 Left shows the zig-zag scanning order; right shows the proposed scanning order with horizontal-frequency major direction.

## VI. EXPERIMENT AND DISCUSSION

In this section, the effectiveness of the proposed DCTNet and PCANet is evaluated on a number of benchmark face datasets namely AR [17], FERET-I ('b' subset) and FERET-II ('fa', 'fb', 'fc', 'dup I', 'dup II' subset)[16].

To have a fair comparison, the PCANet filter learned from Multi-PIE dataset [18] consists of 337 subjects with around 100,000 images (shared by the PCANet's author) is used in the experiment, denoted as **PCANet-A**. The filter is learned for a 2 layers PCANet with filter size $k = 5 \times 5$ for each layer and the number of filter for each layer is $P_1 = P_2 = 8$. We also learn our own PCA filters with the same parameter from gallery for each dataset separately and it is denoted as **PCANet-B**. Lastly, DCTNet with the same parameter is also used (ie, 2D DCT basis of size $5 \times 5$ with 8 bases for each layer). In other words, the experiment is conducted with 3 types of filters – filter learned from external dataset, filter learned from gallery and precomputed filters obtained from 2D DCT. All networks examined are restricted to two layers as we find that the network with more than two layers does not offer significant performance gain whereas incurs higher computation load.

In addition, to evaluate the effectiveness of the proposed histogram TR normalization technique, each experiment is conducted with the presence and absence of the proposed method. Finally, Nearest Neighbor classifier with cosine distance is used for all experiments.

### A. Evaluation on AR Dataset

TABLE I
AR DATASET RECOGNITION RATES (%)

| TR Norm. | Method | Expres. | Illum. | Occlus. | Avg |
|---|---|---|---|---|---|
| No | PCANet-A | **95.960** | 100 | 98.232 | **98.064** |
|  | PCANet-B | 94.276 | 100 | 97.896 | 97.391 |
|  | DCTNet | 94.108 | 100 | 97.643 | 97.250 |
| Yes | PCANet-A | **98.148** | 100 | 99.074 | **99.074** |
|  | PCANet-B | 97.811 | 100 | 99.158 | 98.990 |
|  | DCTNet | 97.811 | 100 | **99.242** | 99.018 |

AR dataset [17] contains 126 subjects with over 4000 images. It is composed of frontal faces with different facial expression, illumination variations and occlusions (sunglasses and scarf). In the experiment, subset of 50 male subjects and 50 female subjects are used. Each image is converted to gray scale and cropped to $165 \times 120$. For gallery, 2 frontal faces with neutral facial expression of each subject are selected and the rest are used as probes which are divided into 3 groups (ie, expression, illumination and occlusion). For all networks, the size of block-wise histogram is set to $20 \times 20$ and the dimension of the final feature vector is reduced to 150 with WPCA.

TABLE II
FERET-I RECOGNITION RATES (%)

| TR Norm. | Method | Bc | Bd | Be | Bf | Bg | Bh | Avg |
|---|---|---|---|---|---|---|---|---|
| No | PCANet-A | 51.5 | 91.0 | 99.0 | 99.5 | 93.0 | 51.5 | 80.92 |
| | PCANet-B | 62.0 | 92.5 | **100** | **100** | 95.5 | 55.5 | 84.25 |
| | DCTNet | **70.5** | **97.0** | 99.5 | **100** | **96.0** | **73.0** | **89.33** |
| Yes | PCANet-A | 82.0 | 97.0 | 100 | 100 | 98.5 | 76.0 | 92.25 |
| | PCANet-B | **88.5** | **99.5** | 100 | 100 | 99.5 | **86.0** | **95.58** |
| | DCTNet | 85.5 | 98.5 | 100 | 100 | 99.5 | 85.0 | 94.75 |

Table I reports the performance of each method. It is observed that all methods are insensitive to illumination variations and robust against facial expression variation and occlusions. DCTNet filter without DC component and PCANet mean removal for each patch make them robust against various lighting conditions.

Big block-wise histogram block size covers bigger area of each face region make it robust against various local deformation such as facial expression variation as reported in [9]. The block-wise histogram that encodes pdf of each face region could be the reason that makes it robust against occlusions. In other words, occluded region yields very different block-wise histogram from all subjects in the gallery at the same region yields low score and is somehow ignored during the match. Another explanation as discussed in [9] could be that the selection of frequency band of PCA and 2D DCT basis as filters, which is based on human facial characteristic as described in section V, leads to low response of the occluded region that does not fall within the frequency bands.

Apart from that, the presence of the proposed TR normalization is observed to boost the performance for both expression and occlusions probe sets. As the gallery set only contains frontal faces with neutral facial expression, the encoded pdf of each block-wise histogram may fit the neutral facial expression well that does not cater expression changes. It shows its advantage over probe set that has different probability distribution from gallery set.

### B. Evaluation on FERET-I

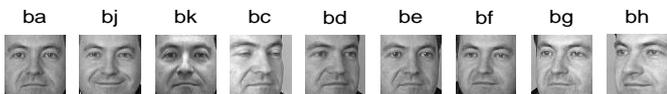

Fig. 6 Samples of FERET 'b' subset

This dataset is the 'b' subset of FERET dataset [16]. It contains 200 subjects with total of 1800 images. Each image is aligned with eyes and mouth coordinate and cropped to $64 \times 64$. The protocol used by [19] is adopted in the experiment, which uses frontal faces with expression and illumination variations (ie, 'ba', 'bj' and 'bk') as gallery set, and non-frontal viewing subset (ie, 'bc', 'bd', 'be', 'bf', 'bg', and 'bh') with pose angle range from +40 to -40 degree are used as probe set. The size block-wise histogram is set to $16 \times 16$ and the final feature vector is used without dimension reduction.

Table II shows that with the absence of histogram TR normalization, the proposed DCTNet has the best performance. Big performance difference between DCTNet and PCANet is observed on probe set with pose angle +40 and -40 (Bc and Bh respectively). Big pose angle in the probe set leads to very different pdf from the training data used by PCANets which only contains frontal face. The learning-free DCTNet that does not rely on training data may be the reason that makes it extracts more generic feature rather than feature that is bound to a specific feature pdf learned from training data.

Furthermore, a surprising huge performance boost is observed on learning based PCANet when TR normalization is applied. The robustness against outliers contributed from the tied-rank as used in Spearman's rank correlation may be one of the reasons to the gain. Moreover, the idea of evenly distributed feature seems contribute to the performance boost too. The square-root operation that compresses large value more and intra-normalization on block-wise histogram that make the resulting histogram more evenly distributed. Here we see that, the advantage of both tied-rank and evenly distributed features make the resulting block-wise histogram be robust when gallery set and probe set have very different pdf.

### C. Evaluation on FERET-II

TABLE III
FERET-II RECOGNITION RATES (%)

| TR Norm. | Method | Fb | Fc | Dup-I | Dup-II | Avg |
|---|---|---|---|---|---|---|
| No | PCANet-A | **99.25** | 100 | **94.46** | **93.16** | **96.72** |
| | PCANet-B | **99.25** | 100 | 93.49 | 91.45 | 96.05 |
| | DCTNet | 99.08 | 100 | 93.35 | 91.45 | 95.97 |
| Yes | PCANet-A | 99.33 | 100 | 94.88 | **94.44** | 97.16 |
| | PCANet-B | 99.58 | 100 | 95.15 | 93.59 | 97.08 |
| | DCTNet | **99.67** | 100 | **95.57** | 94.02 | **97.32** |

Lastly, with the same FERET dataset [16] but different protocol, subset 'fa', 'fb', 'fc', 'dup-I' and 'dup-II' are used in this experiment. Where 'fa' is regular facial expression, 'fb' is different facial expression, 'fc' is face with illumination variation, 'dup-I' probe images were taken between 0 to 1031 days after the gallery match and 'dup-II' probe images were taken at least 18 months after the gallery match which is also a subset of 'dup-I'. In this experiment, we use gray scale images with each cropped to 128×128. Finally 'fa' is used as gallery set and the rest are used as probe sets. For this dataset, the block-wise histogram size is set to 16×16 and the final feature vector is reduced to 1000 dimension with WPCA.

The experiment results as given in Table III shows that DCTNet without TR normalization has the worst performance among other methods. However, with the presence of TR normalization, DCTNet has the overall best recognition rates. Once again the histogram normalization technique consistently boosts the performances of all methods.

### D. Comparison with other methods

To compare the performance of the proposed method with other state-of-the-arts we compile the result of FERET-II in Table IV. The learning free DCTNet achieves the state-of-the-art accuracy with average of 97.32%. Note that, PCANet-2 [9] and PCANet-A use the PCA filter shared by the author which is learned from Multi-PIE dataset. PCANet-2 uses cropped FERET-II image of size 150 × 90 pixels and 15 × 15 block-wise histogram while PCANet-A uses cropped image of size 128 × 128 and 16 × 16 block-wise histogram. With the same dataset used by PCANet-2 we expect some performance gain in DCTNet.

TABLE IV
FERET-II RECOGNITION RATES (%) WITH OTHER METHODS

| Method | Fb | Fc | Dup-I | Dup-II | Avg |
|---|---|---|---|---|---|
| LBP [12] | 93.00 | 51.00 | 61.00 | 50.00 | 63.75 |
| DMMA [20] | 98.10 | 98.50 | 81.60 | 83.20 | 90.35 |
| G-LBP [21] | 98.00 | 98.00 | 90.00 | 85.00 | 92.75 |
| WPCA-POEM [22] | 99.60 | 99.50 | 88.80 | 85.00 | 93.23 |
| G-LQP [23] | **99.90** | **100** | 93.20 | 91.00 | 96.03 |
| LGBP-LGXP [24] | 99.00 | 99.00 | 94.00 | 93.00 | 96.25 |
| sPOEM+POD [25] | 99.70 | **100** | 94.90 | 94.00 | 97.15 |
| GOM [26] | **99.90** | **100** | **95.70** | 93.10 | 97.18 |
| PCANet-2 [9] | 99.58 | **100** | 95.43 | **94.02** | 97.26 |
| PCANet-A | 99.25 | **100** | 94.46 | 93.16 | 96.72 |
| DCTNet | 99.67 | **100** | 95.57 | **94.02** | **97.32** |

### VII. DISCUSSION AND CONCLUSIONS

In this paper, the proposed learning free DCTNet gives us a different perspective of the filters learned by PCANet. The nature of image local correlation characteristic that can be modeled with stationary first order Markov process with the assumption that the neighboring pixels are highly correlated leading us to a much simple learning-free convolutional network. The relationship of frequency and variance of PCA and 2D DCT leads us to rank the 2D DCT basis importance from the lowest frequency as filter selection and it is demonstrated on various face datasets to work very well. On the down side, DCTNet may not work well if the nature of input image does not follow the high local correlation assumption such as image that contains high spectral activity and fine details like texture images. Such image data may need different DCT basis selection schemes.

On the bright side, as long as the input image meets the model assumption which happened to be the nature of most natural images, makes the learning-free DCTNet stand out. PCANet on the other hand that relies on training data to learn the filters may over fit especially if the probe set distribution is far deviated from the training set as observed in FERET 'b' subset experiment without histogram TR normalization.

In conjunction with the proposed histogram TR normalization technique, DCTNet contributes a huge performance gain as observed in FERET 'b' subset experiment where the frontal face training data and probe with large pose angle may have very different distribution. AR 'expression' subset and FERET aging (dup-I and dup-II) subset that have local facial deformations are shown to have some gain in performance too. The proposed histogram TR normalization method can also be seen as a post-processing method to regulate the extracted block-wise histogram from representing the subject with the gallery specific distribution.

To conclude, despite learning free, the remarkable performance from extensive face recognition experiments, which comprise of illumination variation, facial expression variation, occlusions, pose and time span endorses the capability of DCTNet. Indeed, each component of the network which play different roles in extracting invariant and discriminative feature is important for DCTNet to achieve good performance.

ACKNOWLEDGMENT

This work was supported by Basic Science Research Program through the National Research Foundation of Korea (NRF) funded by Ministry of Science, ICT and Future Planning (2013006574) and Institute of BioMed-IT, Energy-IT and SmartIT Technology (BEST), a Brain Korea 21 Plus Program, Yonsei University.